\documentclass[graybox]{svmult}

\usepackage{mathptmx}       
\usepackage{helvet}         
\usepackage{courier}        
\usepackage{type1cm}        
                            
\usepackage{makeidx}         
\usepackage{graphicx}        
                             
\graphicspath{{Figures/}}

\usepackage{multicol}        
\usepackage[bottom]{footmisc}
\usepackage[utf8]{inputenc}

\usepackage[numbers,sort&compress,sectionbib]{natbib}

\usepackage{eucal}
\usepackage{amsmath}
\usepackage{amsfonts}

\newcommand{\etal}{\textit{et al.~}}

\makeindex

\begin{document}

\title*{Generative Models for Automatic Chemical Design}

\author{Daniel Schwalbe-Koda and Rafael G\'omez-Bombarelli}

\institute{Daniel Schwalbe-Koda \at Department of Materials Science and Engineering, Massachusetts Institute of Technology, \email{dskoda@mit.edu}
\and Rafael G\'omez-Bombarelli \at Department of Materials Science and Engineering, Massachusetts Institute of Technology, \email{rafagb@mit.edu}}

\maketitle

\abstract*{Materials discovery is decisive for tackling urgent challenges related to energy, the environment,  health care and many others. In chemistry, conventional methodologies for innovation usually rely on expensive and incremental strategies to optimize properties from molecular structures. On the other hand, inverse approaches map properties to structures, thus expediting the design of novel useful compounds. In this chapter, we examine the way in which current deep generative models are addressing the inverse chemical discovery paradigm. We begin by revisiting early inverse design algorithms. Then, we introduce generative models for molecular systems and categorize them according to their architecture and molecular representation. Using this classification, we review the evolution and performance of important molecular generation schemes reported in the literature. Finally, we conclude highlighting the prospects and challenges of generative models as cutting edge tools in materials discovery.}

\abstract{Materials discovery is decisive for tackling urgent challenges related to energy, the environment,  health care and many others. In chemistry, conventional methodologies for innovation usually rely on expensive and incremental strategies to optimize properties from molecular structures. On the other hand, inverse approaches map properties to structures, thus expediting the design of novel useful compounds. In this chapter, we examine the way in which current deep generative models are addressing the inverse chemical discovery paradigm. We begin by revisiting early inverse design algorithms. Then, we introduce generative models for molecular systems and categorize them according to their architecture and molecular representation. Using this classification, we review the evolution and performance of important molecular generation schemes reported in the literature. Finally, we conclude highlighting the prospects and challenges of generative models as cutting edge tools in materials discovery.}

\section{Introduction}

Innovation in materials is the key driver for many recent technological advances. From clean energy \cite{Tabor2018Accelerating} to the aerospace industry \cite{Gibson2010review} or drug discovery \cite{Chen2018rise}, research in chemical and materials science is constantly pushed forward to develop compounds and formulae with novel applications, lower cost and better performance. Conventional methods for the discovery of new materials start from a well-defined set of substances from which properties of interest are derived. Then, intensive research on the relationship between structures and properties is performed. The gained insights from this procedure lead to incremental improvements in the compounds and the cycle is restarted with a new search space to be explored. This trial-and-error approach to innovation often leads to costly and incremental steps towards the development of new technologies and in occasion relies on serendipity for leap progress. Materials development may require billions of dollars in investments \cite{DiMasi2016Innovation} and up to 20 years to be deployed to the market \cite{DiMasi2016Innovation,Tabor2018Accelerating}.

Despite the challenges associated with such direct approaches, they have not prevented data-driven discovery of materials from happening. High-throughput materials screening \cite{Shoichet2004Virtual,Greeley2006Computational,Alapati2006Identification,Setyawan2011High,Subramaniam2008Virtual,Armiento2011Screening,Jain2011high,Curtarolo2013high,Pyzer-Knapp2015What,Gomez-Bombarelli2016Design} and data mining \cite{Morgan2004High,Ortiz2009Data,Yu2012Identification,Yang2012search,Lin2012silico,Mounet2018Two} have been responsible for several breakthroughs in the last two decades \cite{Potyrailo2011Combinatorial,Jain2016Computational}, leading to the establishment of the Materials Genome Initiative \cite{NSTC2011Materials} and multiple collaborative projects around the world build around databases and analysis pipelines \cite{Curtarolo2012AFLOWLIB.ORG,Calderon2015AFLOW,Jain2013Commentary,Saal2013Materials}. Automated, scalable approaches leverage from data sets in the thousands to millions of simulations to offer a cornucopia of insights on materials composition, structure and synthesis.

Developing materials with the inverse perspective departs from these traditional methods. Instead of exhaustively deriving properties from structures, the performance parameters are chosen beforehand and unknown materials satisfying these requirements are inferred. Hence, innovation in this setting is achieved by reverting the mapping between structures and their properties. Unfortunately, this approach is even harder than the conventional one. Inverting a given Hamiltonian is not a well-defined problem, and the absence of a systematic exploratory methodology may result in delays, or outright failure, of the discovery cycle of materials \cite{Sanchez-Lengeling2018Inverse}. Furthermore, another major obstacle to the design of arbitrary compounds is the dimensionality of the missing data for known and unknown compounds \cite{Zunger2018Inverse}. As an example, the breadth of accessible drug-like molecules can be on the order of $10^{60}$ \cite{Polishchuk2013Estimation,Virshup2013Stochastic}, rendering manual searches or enumerations through the chemical space an intractable problem. In addition, molecules and crystal structures are discrete objects, which hinders automated optimization, and computer-generated candidates must follow a series of hard (valence rules, thermal stability) and soft (synthetic accessibility, cost, safety) constraints that may be difficult to state in explicit form. As the inverse chemical design holds great promise for economic, environmental and societal progress, one can ask how to rationalize the exploration of unknown substances and accelerate the discovery of new materials.

\subsection{Early inverse design strategies for materials}

The inverse chemical design is usually posed as an optimization problem in which molecular properties are extremized with respect to given parameters \cite{Joback1989Designing}. This concept splits the inverse design problem in two parts: (i) efficiently sampling materials from an enormous configuration space, and (ii) searching for global maxima in their properties \cite{Kuhn1996Inverse} corresponding to minima in their potential energy surface \cite{Wales1999Global,Schoen2001Determination}. Early approaches towards the inverse materials design used chemical intuition to address (i), narrowing down and navigating the space of structures under investigation with probabilistic methods \cite{Gani1983Molecular,Marder1991Approaches,Holmblad1996Designing,Kuhn1996Inverse,Sigmund1997Design,Wolverton1997Invertible}. Nevertheless, even constrained spaces can be too large to be exhaustively enumerated. Especially in the absence of an efficient exploratory policy, this discovery process demands considerable computational resources and time. Several different strategies are required to simultaneously navigate the chemical space and evaluate the properties of the materials under investigation.

Monte Carlo methods resort to statistical sampling to avoid enumerating a space of interest. When combined with simulated annealing  \cite{Metropolis1953Equation}, for example, they become adequate to locate extrema within property spaces. In physics, reverse Monte Carlo methods have long been developed to determine structural information from experimental data \cite{Kaplow1968Atomic,Gerold1987determination,McGreevy1988Reverse}. However, the popularization of similar methods to \textit{de novo} design of materials is more recent. Wolverton \etal \cite{Wolverton1997Invertible} employed such methods to aid the design of alloys and avoid expensive enumeration of compositions and Franceschetti and Zunger \cite{Franceschetti1999inverse} improved the idea to design Al$_x$Ga$_{1-x}$As and Ga$_x$In$_{1-x}$P superlattices with a tailored band gaps. They started with configurations sampled using Monte Carlo, relaxed the atomic positions using valence-force-field methods and calculated their band gap by fast diagonalization of pseudopotential hamiltonians. Through this practical process, they predicted superlattices with optimal band gaps after analyzing less than $10^4$ compounds among $\sim 10^{14}$ structures \cite{Franceschetti1999inverse}.

Other popular techniques for multidimensional optimization that also involve a stochastic component are genetic algorithms (GAs) \cite{Holland1992Adaptation}. Based on evolution principles, GAs refine specific parameters of a population that improve a targeted property. In materials design, GAs have been vastly employed in the inverse design of small molecules \cite{Judson1993Conformational,Glen1995genetic}, polymers \cite{Venkatasubramanian1994Computer,Venkatasubramanian1995Evolutionary}, drugs \cite{Parrill1996Evolutionary,Schneider2000De}, biomolecules \cite{Gordon1999Branch,Reetz2004Asymmetric}, catalysts \cite{Wolf2000evolutionary}, alloys \cite{Johannesson2002Combined,Dudiy2006Searching}, semiconductors \cite{Piquini2008Band,dAvezac2012Genetic,Zhang2013Genetic}, and photovoltaic materials \cite{Yu2012Inverse}. Furthermore, evolution-inspired approaches have been used as a general modeling tool to predict stable structures \cite{Brodmeier1994Application,Woodley1999prediction,Glass2006USPEXEvolutionary,Oganov2006Crystal,Froemming2009Optimizing,Vilhelmsen2014genetic} and Hamiltonian parameters \cite{Hart2005Evolutionary,Blum2005Using}. Many more applications of GAs in materials design are still being demonstrated after decades of its inception \cite{Virshup2013Stochastic,Rupakheti2015Strategy,Reymond2015Chemical,Le2016Discovery,Jennings2019Genetic}.

Monte Carlo and evolutionary algorithms are interpretable and often produce powerful implementations. The combination of sampling and optimization is a great improvement over random searches or full enumeration of a chemical space. Nonetheless, they still correspond to discrete optimization techniques in a combinatorial space, and require individual evaluation of their properties at every step. This discrete form hinders chemical interpolations and the definition of property gradients during optimization processes, thus retaining a flavor of ``trial-and-error'' in the computational design of materials, rather than an invertible structure-property mapping. One of the first attempts to use a continuous representation on the molecular design was performed by Kuhn and Beratan \cite{Kuhn1996Inverse}. The authors varied coefficients in linear combination of atomic orbitals while keeping the energy eigenvalues fixed to optimize linear chains of atoms. Later, Lilienfeld \etal \cite{Lilienfeld2005Variational} generalized the discrete nature of atoms by approximating atomic numbers by continuous functions and defining property gradients with respect to this ``alchemical potential''. They used this theory to design ligands for proteins \cite{Lilienfeld2005Variational} and tune electronic properties of derivatives of benzene \cite{Marcon2007Tuning}. A similar strategy was proposed by Wang \etal \cite{Wang2006Designing} around the same time. Instead of atomic numbers, a linear combination of atomic potentials was used as a basis for optimizations in property landscapes. Following the bijectiveness between potential and electronic density in the Hohenberg-Kohn theory \cite{Hohenberg1964Inhomogeneous}, nuclei-electrons interaction potentials were employed as quasi-invertible representations of molecules. Potentials resulting from optimizations with property gradients can be later interpolated or approximated by a discrete molecular structure whose atomic coordinates give rise to a similar potential. Over the years, the approach was further refined within the tight-binding framework \cite{Xiao2008Inverse,Balamurugan2008Exploring} and gradient-directed Monte Carlo method \cite{Keinan2007Designing,Hu2008gradient}, its applicability demonstrated in the design of molecules with improved hyperpolarizability \cite{Wang2006Designing,Keinan2007Designing,Xiao2008Inverse} and acidity \cite{Vleeschouwer2012Inverse}.

Despite these promising approaches, many challenges in inverse chemical design remain unsolved. Monte Carlo and genetic algorithms share the complexity of discrete optimization methods over graphs, particularly exacerbated by the rugged property surfaces. They rely on stochastic steps that struggle to capture the interrelated hard and soft constraints of chemical design: converting a single into a double bond may produce a formally valid, but impractical and unacceptable molecule depending on chemical context. On the other hand, a compromise between validity and diversity of the chemical space is difficult to achieve with continuous representations. Lastly, finding optimal points in the 3D potential energy surface that produce a desired output is still not the same as molecular optimization, since the generated ``atom cloud'' may not be a local minimum, stable enough in operating conditions, or synthetically attainable. An ideal inverse chemical design tool would offer the best of the two worlds: an efficient way to sample valid and acceptable regions of the chemical space; a fast method to calculate properties from given structures; a differentiable representation for a wide spectrum of materials; and the capacity to optimize them using property gradients. Furthermore, it should operate on the manifold of synthetically accessible, stable compounds. This is where modern machine learning (ML) algorithms come into play.

\subsection{Deep learning and generative models}

Deep learning (DL) is emerging as a promising tool to address the inverse design of many different applications. Particularly through generative models, algorithms in DL push forward how machines understand real data. Roughly speaking, the role of a generative model is to capture the underlying rules of a data distribution. Given a collection of (training) data points $\{X_i\}$ in a space $\mathcal{X}$, a model is trained to match the data distribution $P_X$ by means of a generative process $P_G$ in such a way that generated data $Y \sim P_G$ resembles the real data $X \sim P_X$. Earlier generative models such as Boltzmann Machines \cite{Hinton1986Learning,Hinton1983Optimal}, Restricted Boltzmann Machines \cite{Smolensky1986Information}, Deep Belief Networks \cite{Hinton2006Fast} or Deep Boltzmann Machines \cite{Salakhutdinov2009Deep} were the first to tackle the problem of learning probability distributions based on training examples. Their lack of flexibility, tractability and generalizing ability, however, rendered them obsolete in favor of more modern ones \cite{Goodfellow2016Deep}.

Current generative models have been successful in learning and generating novel data from different types of real-world examples. Deep neural networks trained on image datasets are able to produce realistic-looking house interiors, animals, buildings, objects and human faces \cite{Karras2017Progressive,Goodfellow2014Generative}, as well as embed pictures with artistic style \cite{Gatys2015Neural} or enhance it with super-resolution \cite{Ledig2016Photo}. Other examples include convincing text \cite{Bowman2015Generating,Xu2015Show}, music \cite{Mehri2016SampleRNN}, voices \cite{Oord2016WaveNet} and videos \cite{Vondrick2016Generating} synthesized by such networks. Most interesting is the creation of novel data conditioned on latent features, which allows tuning models with vector and arithmetic operations in a property space \cite{Radford2015Unsupervised,Engel2017Latent}. The adaptable architectures of these models also enable straightforward training procedures based on backpropagation \cite{LeCun2015Deep}. Within the DL framework, a proper loss function drives gradients so that the generative model, typically parameterized by a neural network, learns to minimize the distance between the two distributions.

\begin{figure}[tb]
\centering
\includegraphics[width=\linewidth]{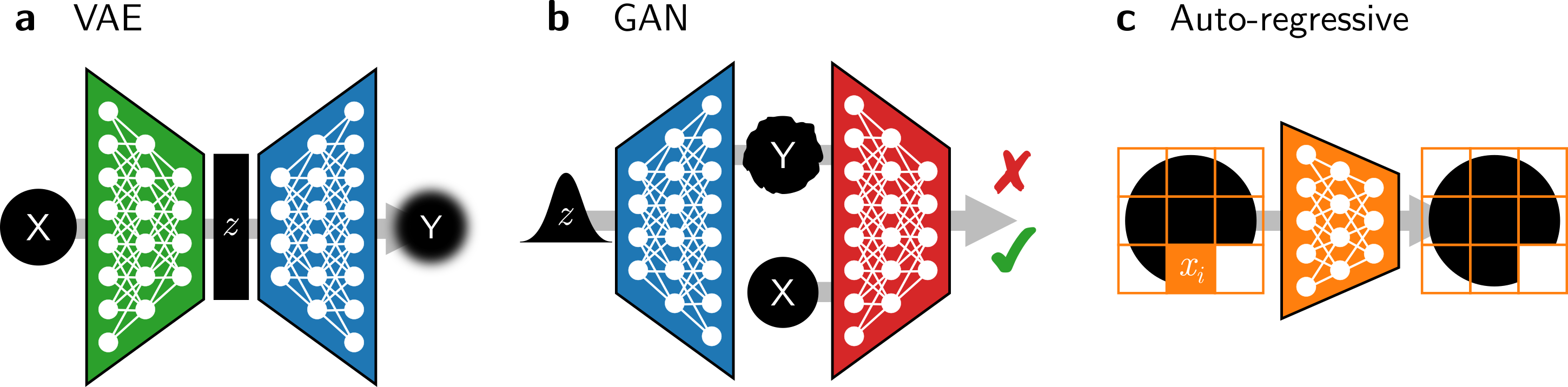}
\caption{Schematic diagrams for three popular generative models: (a) VAE, (b) GAN, and (c) auto-regressive.}
\label{fig:gen_models}
\end{figure}

Among the popular architectures for generating data from deep neural networks, the Variational Auto-Encoder (VAE) \cite{Kingma2013Auto} is a particularly robust architecture. It couples inference and generation by mapping data to a manifold conditioned to implicit data descriptors. To do so, the model is trained to learn the identity function while constrained by a dimensional bottleneck called latent space (see Fig. \ref{fig:gen_models}a). In this scheme, data is first encoded to a probability distribution $Q_\phi (\mathbf{z}|X)$ matching a given prior distribution $P_z(\mathbf{z})$, where $\mathbf{z}$ is called latent vector. Then, a sample from the latent space is reconstructed with the generative algorithm $P_\theta (X|\mathbf{z})$. In the VAE \cite{Kingma2013Auto}, outcomes of both processes are parameterized by $\phi$ and $\theta$ to maximize a lower bound for the log-likelihood of the output with respect to the input data distribution. The VAE objective is, therefore,

\begin{equation}
\mathcal{L}(\theta, \phi) = - D_{KL} \left( Q_\phi(\mathbf{z}|X) || P_z(\mathbf{z}) \right) + \mathbb{E}_{z \sim Q_\phi} \left[ \log P_\theta (X|\mathbf{z})\right].
\end{equation}

\noindent The encoder is regularized with a divergence term $D_{KL}$, while the decoder is penalized by a reconstruction error $\log P_\theta (X|\mathbf{z})$, usually in the form of mean-squared or cross entropy losses. This maximization can then be performed by stochastic gradient ascent.

The probabilistic nature of VAE manifolds approximately accounts for many complex interactions between data points. Although functional in many cases, the modeled data distribution does not always converge to real data distributions \cite{Arjovsky2017Wasserstein}. Furthermore, Kullback-Leibler or Jensen-Shannon divergences cannot be analytically computed for an arbitrary prior, and most works are restricted to Gaussian distributions. Avoiding high-variance methods to determine this regularizing term is also an important concern. Recently, this limitation was simplified by employing the Wasserstein distance as a penalty for the encoder regularization \cite{Arjovsky2017Wasserstein,Tolstikhin2017Wasserstein}. As a result, richer latent representations are computed more efficiently within Wasserstein Auto-Encoders, resulting in disentanglement, latent shaping, and improved reconstruction \cite{Rubenstein2018Latent,Arjovsky2017Wasserstein,Tolstikhin2017Wasserstein}.

Another approach to generative models are the Generative Adversarial Networks (GANs) \cite{Goodfellow2014Generative}. Recognized by their sharp reconstructions, GANs are constructed by making two neural networks compete against each other until a Nash equilibrium is found. One of the networks is a deterministic generative model. It applies a non-linear set of transformations to a prior probability distribution $P_z$ in order to match the real data distribution $P_X$. Interestingly, the generator (or actor) only receives the prior distribution as input, and has no contact with the real data whatsoever. It can only be trained through a second network, called discriminator or critic. The latter tries to distinguish real data $X \sim P_X$ from fake data $Y = G(\mathbf{z}) \sim P_G$, as depicted in Fig. \ref{fig:gen_models}b. The objective of the critic is to perfectly distinguish between $P_X$ and $P_G$, thus maximizing the prediction accuracy. On the other hand, the generator tries to fool the discriminator by creating data points that look like real data points,  minimizing the prediction accuracy of the critic. Consequently, the complete GAN objective is written as \cite{Goodfellow2014Generative}

\begin{equation}
\min_G \max_D V(D, G) = \mathbb{E}_{X \sim P_X} \left[ \log D(X) \right] + \mathbb{E}_{\mathbf{z} \sim P_z} \left[ \log \left(1 - D(G(\mathbf{z}))\right) \right].
\end{equation}

Despite the impressive results from GANs, their training process is highly unstable. The min-max problem requires a well-balanced training from both networks to ensure non-vanishing gradients and convergence to a successful model. Furthermore, GANs do not reward diversity of generated samples and the system is prone to mode collapse. There is no reason why the generated distribution $P_G$ should have the same support of the original data $P_X$, and the actor produces only a handful of different examples which are realistic enough. This does not happen for the VAE, since the log-likelihood term gives an infinite loss for a generated data distribution with a disjoint support with respect to the original data distribution. Several different architectures have been proposed to address these issues among GANs \cite{Mirza2014Conditional,Chen2016InfoGAN,Arjovsky2017Wasserstein,Che2016Mode,Odena2016Conditional,Mao2016Least,Hjelm2017Boundary,Zhao2016Energy,Nowozin2016f,Donahue2016Adversarial,Berthelot2017BEGAN,Gulrajani2017Improved,Yi2017DualGAN}. Although many of them may be equivalent to a certain extent \cite{Lucic2017Are}, steady progress is being made in this area, especially through more complex ways of approximating data distributions, such as with f-divergence \cite{Nowozin2016f} or optimal transport \cite{Arjovsky2017Wasserstein,Gulrajani2017Improved,Berthelot2017BEGAN}.

Other models such as the auto-regressive PixelRNN \cite{Oord2016Pixel} and PixelCNN \cite{Oord2016Conditional,Salimans2017PixelCNN++} have also been successful as generators of images \cite{Oord2016Pixel,Oord2016Conditional,Salimans2017PixelCNN++}, video \cite{Kalchbrenner2016Video}, text \cite{Kalchbrenner2016Neural} and sound \cite{Oord2016WaveNet}. Differently from VAE and GANs, these models approximate the data distribution by a tractable factorization $P_X$. For example, in an $n \times n$ image, the generative model $P(X)$ is written as \cite{Oord2016Pixel}

\begin{equation}
P(X) = \prod_{i=1}^{n^2} P\left(x_i | x_1, \ldots, x_{i-1} \right),
\end{equation}

\noindent where each $x_i$ is a pixel generated by the model (see Fig. \ref{fig:gen_models}c). These models with explicit distributions yield samples with very good negative log-likelihood and diversity \cite{Oord2016Pixel}. The model evaluation is also straightforward, given the explicit computation of $P(X)$. As a drawback, however, these models rely on the sequential generation of data, which is a slow process. A diagram of the architectures of the three generative models here discussed is seen in Fig. \ref{fig:gen_models}.

\subsection{Generative models meet chemical design}

Apart from their numerous aforementioned applications, generative models are also attracting attention in chemistry and materials science. DL is being employed not only for the prediction and identification of properties of molecules, but also to generate new chemical compounds \cite{LeCun2015Deep}. In the context of inverse design, generative models provide benefits such as: generating complex samples from simple probability distributions; providing meaningful latent representations, over which optimizations can be performed; and the ability to perform inference when coupled to supervised models. Therefore, unifying generative models with chemical design is a promising venue to accelerate innovation in chemistry and related fields.

To go beyond the limitations of traditional inverse design strategies, an ideal way to discover new materials should satisfy some requisites \cite{Gomez-Bombarelli2018Automatic}. To be a completely hands-free model, the model should be data-driven, thus avoiding fixed libraries and expensive labeling. It is also desirable that it outputs as many potential molecules as possible under a subset of interest, which means that the model needs a powerful generator coupled with a continuous representation for molecules. Furthermore, such a representation should be interpretable, allowing a correct description of structure-property relationships within molecules. If, additionally, the model is differentiable, it would be possible to optimize certain properties using gradient techniques and, later, look for molecules satisfying such constraints.

The development of such a tool is currently a priority for ML models in chemistry and for the inverse chemical design. It relies primarily on two decisions: which model to use and how to represent a molecule in a computer-friendly way. Following our brief introduction to the early inverse design strategies and main generative models in the literature, we describe which molecular representations are possible.

In quantum mechanics, a molecular system is represented by a wavefunction that is a solution of the Schr\"odinger equation for that particular molecule. To derive most properties of interest, the spatial wavefunction is enough. Computing such a representation, however, is equivalent to solving an (approximate) version of the Schr\"odinger equation itself. Many methods for theoretical chemistry, such as Hartree-Fock \cite{Hartree1928Wave,Fock1930Naherungsmethode} or Density Functional Theory \cite{Hohenberg1964Inhomogeneous,Kohn1965Self}, represent molecules using wavefunctions or electronic densities and obtain other properties from it. Solving quantum chemical calculations is computationally demanding in many cases, though. The idea with many ML methods is not only to avoid these calculations, but also to make a generalizable model that highlight different aspects of chemical intuition. Therefore, we should look for other representations for chemical structures.

\begin{figure}[htb]
\centering
\includegraphics[width=.8\linewidth]{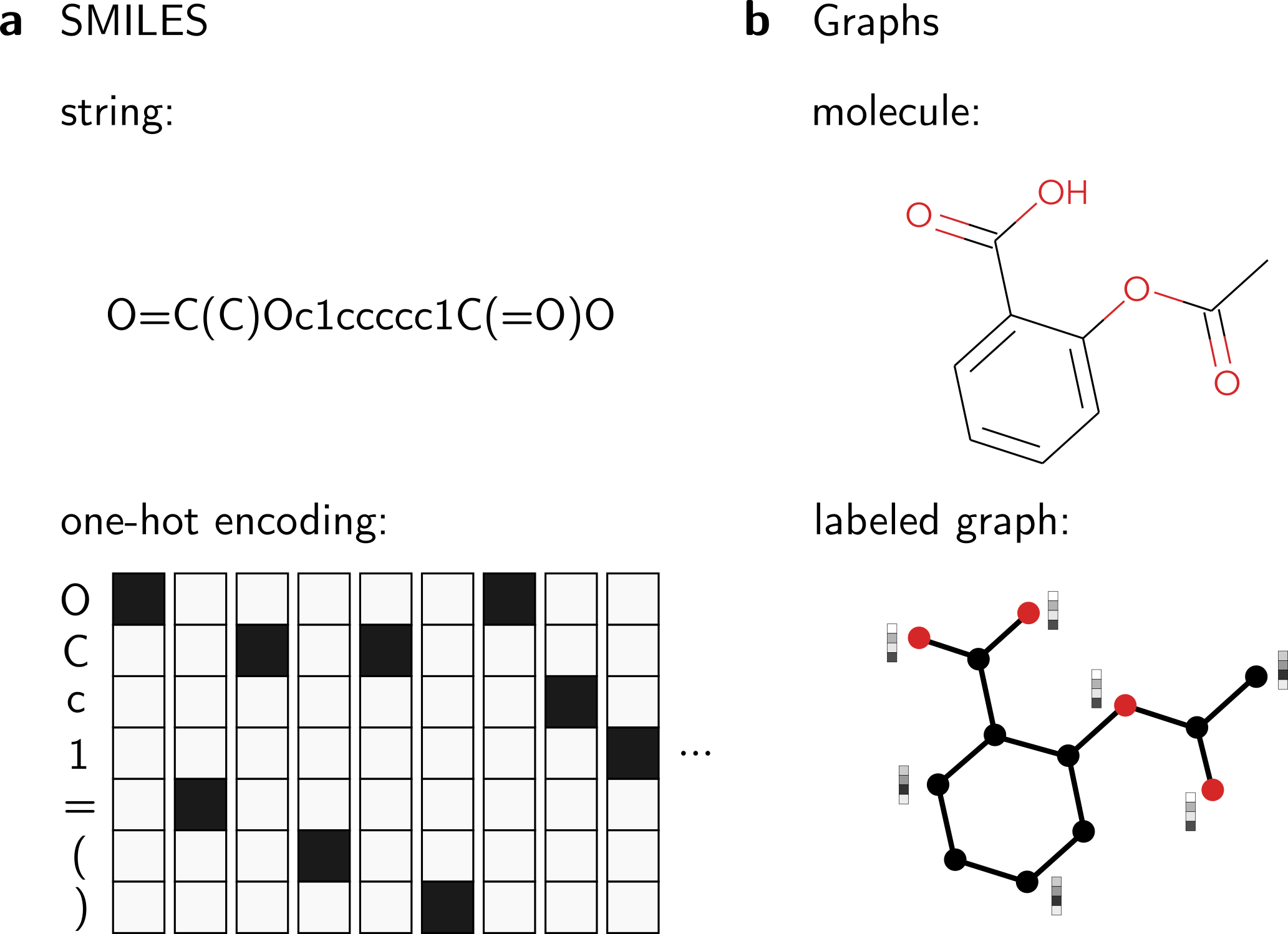}
\caption{Two popular ways of representing a molecule using: (a) SMILES strings converted to one-hot encoding; or (b) a graph derived from the Lewis structure.}
\label{fig:representations}
\end{figure}

Thousands of different descriptors are available for chemical prediction methods \cite{Todeschini2000Handbook}. Several relevant features for ML have demonstrated their capabilities for predicting properties of molecules, such as fingerprints \cite{Rogers2010Extended}, bag-of-bonds \cite{Hansen2015Machine}, Coulomb matrices \cite{Rupp2012Fast}, deep tensor neural networks train on the distance matrix\cite{Schuett2017Quantum}, many-body tensor representation \cite{Huo2017Unified}, SMILES strings \cite{Weininger1988SMILES}, and graphs \cite{Kearnes2016Molecular,Duvenaud2015Convolutional,Gilmer2017Neural}. Not all representations are invertible for human interpretation, however. To teach a generative model how to create a molecule, it may suffice for it to produce a fingerprint, for example. However, how can one map any possible fingerprint to a molecule is an extra step of complexity equivalent to the generation of libraries. This is undesirable in a practical generative model. In this chapter, we focus on two easily interpretable representations, SMILES strings and molecular graphs, and how generative models perform with these representations. Examples of these two forms of writing a molecule are shown in Fig. \ref{fig:representations}.

\section{Chemical generative models}

\subsection{SMILES representation}

SMILES (Simplified Molecular Input Line Entry System) strings have been widely adopted as representation for molecules \cite{Weininger1988SMILES}. Through graph-to-text mapping algorithms, it determines atoms by atomic number and aromaticity, and can capture branching, cycles, ionization, etc. The same molecule can be represented by multiple SMILES strings, and thus a canonical representation is typically chose, although some works leverage non-canonical strings as a data augmentation and regularization strategy. Although SMILES are inferior to the more modern InChI (International Chemical Identifier) representation in their ability to address key challenges in representing molecules as strings such as tautomerism, mesomerism and some forms of isomerism, SMILES follow a much simpler syntax that has proven easier to learn for ML models.

Since SMILES rely on a sequence-based representation, natural language processing (NLP) algorithms in deep learning can be naturally extended to them. This allows the transferability of several architectures from the NLP community to interpret the chemical world. Mostly, these systems make use of recurrent neural networks (RNNs) to condition the generation of the next character on the previous ones, creating arbitrarily long sequences character by character \cite{Goodfellow2016Deep}. The order of the sequence is very relevant to generate a valid molecule, and observation of such restrictions can be typically incorporated in RNNs with long short-term memory cells (LSTM) \cite{Hochreiter1997Long}, gated recurrent units (GRUs) \cite{Chung2014Empirical}, or stack-augmented memory \cite{Popova2018Deep}.

A simple form of generating molecules using only RNN architectures is to extensively train them with valid SMILES from a database of molecules. This requires post-processing analyses, as it resembles traditional library generation. As a proof of concept, Ikebata \etal \cite{Ikebata2017Bayesian} used SMILES strings to design small organic molecules by employing Bayesian sampling with sequential Monte Carlo. Ertl \etal \cite{Ertl2017silico} instead generated molecules using LSTM cells and later employed them in a virtual screening for properties.

Generating libraries, however, is not enough for the automatic discovery of chemical compounds. Asking an RNN-based model to simply create SMILES strings does not improve on the rational exploration of the chemical space. In general, the design of new molecules is also oriented towards certain properties, like solubility, toxicity and drug-likeness \cite{Gomez-Bombarelli2018Automatic}, which are not necessarily incorporated in the training process of RNNs. In order to skew the generation of molecules and better investigate a subset of the chemical space, Segler \etal \cite{Segler2018Generating} used transfer-learning to first train the RNN on a whole dataset of molecules and later fine-tune the model towards the generation of molecules with physico-chemical properties of interest. This two-part approach allows the model to first learn the grammar inherent to SMILES to then create new molecules based only on the most interesting ones. In line with this depth-search, Gupta \etal \cite{Gupta2017Generative} demonstrated the application of transfer learning to grow molecules from fragments. This technique is particularly useful for drug discovery \cite{Chen2018rise, Ching2018Opportunities}, in which the search of the chemical space usually begins from a known substructure with certain desired functionalities.

Recently, the usage of reinforcement learning (RL) to generate molecules with certain properties became popular among generative models. Since the representation of a molecule using SMILES requires the generator to output a sequence of characters, each decision can be considered as an action. The successful completion of a valid SMILES string is associated with a reward, for example, and undesired features in the sequence are penalized. Jaques \etal \cite{Jaques2017Sequence} used RL to impose a structure on sequence generation, avoiding repeating patterns not only in SMILES strings but also in text and music. By penalizing large rings, short sequences of characters and long, monotonous carbon chains, they were able to increase the number of valid molecules their model produced. Olivecrona \etal \cite{Olivecrona2017Molecular} demonstrated the usage of augmented episodic likelihood and traditional policy gradient methods to tune the generation of molecules from an RNN. Their method achieved 94\% of validity on generating molecules sampled from a prior distribution. It was also taught to avoid functional groups containing sulfur and to generate structures similar to a given structure or with certain target activities. Similarly, Popova \etal \cite{Popova2018Deep} designed molecules for drugs using a stack-augmented RNN. It demonstrated improved capacity to capture the grammar of SMILES while using RL to tune their synthetic accessibility, solubility, inhibition and other properties.

As the degree of abstraction grows in the molecule design, more complex generative models are proposed to explore the chemical space. VAEs, for example, can include a direct mapping between structures and properties and vice-versa. Its joint training with an encoder and a decoder is capable of approximating very complex data distributions using a real-valued and compressed representation, which is essential for improving the search for chemical compounds. Since the latent space is meaningful, the generator learns to associate patterns in the latent space with properties of the real data. After both the encoding and the decoding networks are jointly trained, the generative model can be decoupled from the inference step and latent variables then become the field for exploration. Therefore, VAEs map the original chemical space to a continuous, differentiable space conveying all the information about the original molecules, over which optimization can be performed. Additionally, conditional generation of molecules based on properties is made possible without hand-made constraints in SMILES, semi-supervised methods can be used to tune the model with relevant properties. This approach is closer to the model of an ideal, automatic, chemical generative model as discussed earlier.

\begin{figure}[tb]
\centering
\includegraphics[width=.7\linewidth]{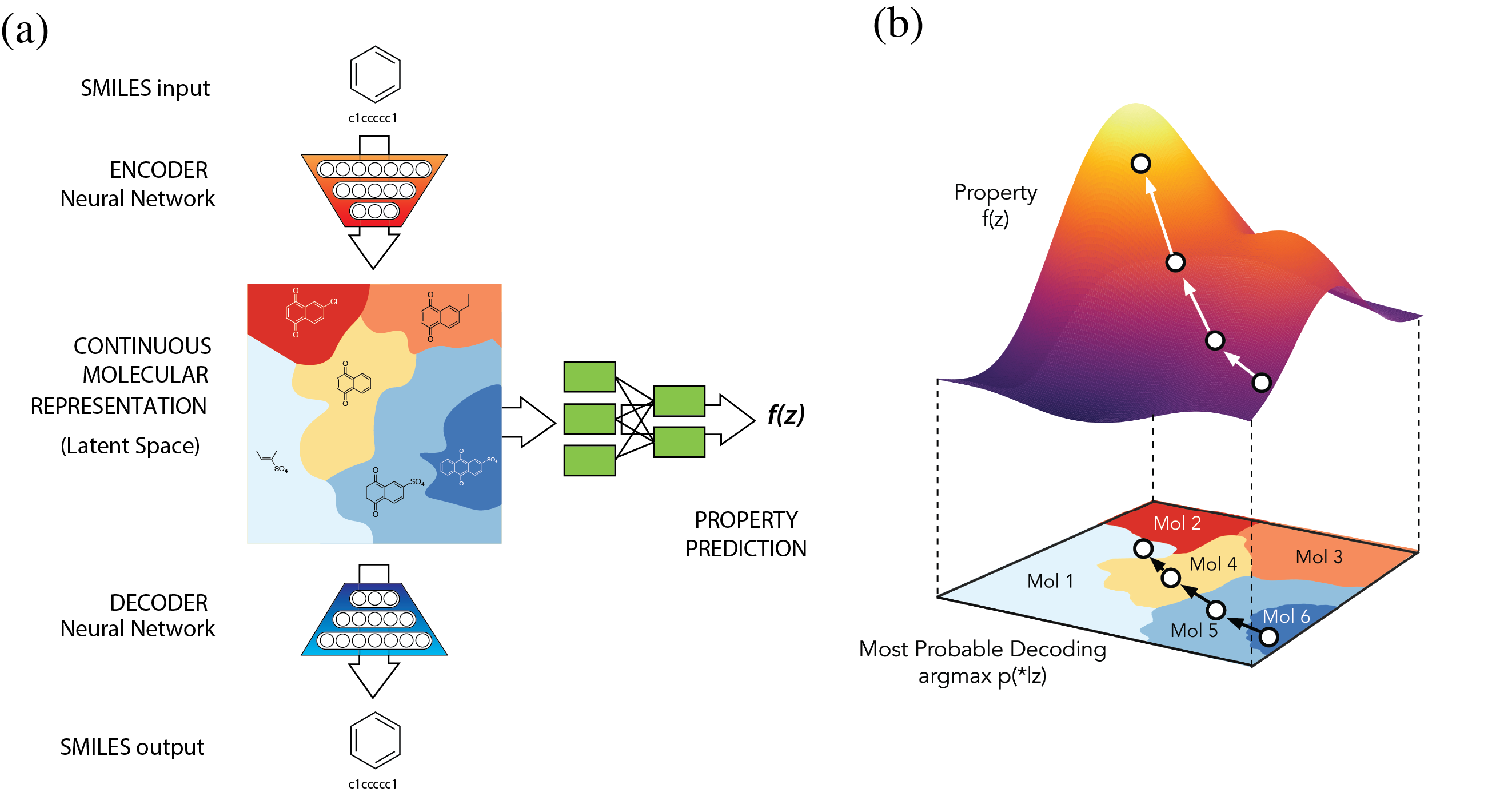}
\caption{Variational Auto-Encoder for chemical design. The architecture in (a) allows for property optimization in the latent space, as depicted in (b). Figure reproduced from \cite{Gomez-Bombarelli2018Automatic}.}
\label{fig:vae_smiles}
\end{figure}

Constructed over RNNs as both encoder and decoder, G\'omez-Bombarelli \etal \cite{Gomez-Bombarelli2018Automatic} trained a VAE on prediction and reconstruction tasks for molecules extracted from the QM9 and ZINC datasets. The latent space allowed not only sampling of molecules but also interpolations, reconstruction, and optimization using a Gaussian process predictor trained on the latent space (Fig. \ref{fig:vae_smiles}).  Kang and Cho \cite{Kang2018Conditional} used partial annotation on molecules to train a semi-supervised VAE to decrease the error for property prediction and to generate molecules conditioned on targets. It can also be enhanced in combination with other dimensionality reduction algorithms \cite{Sattarov2019De}. Within the chemical world, VAEs based on sequences also show promise for investigating proteins \cite{Sinai2017Variational}, learning chemical interactions between molecules \cite{Kwon2017DeepCCI}, designing organic light-emitting diodes \cite{Kim2018Deep} and generating ligands \cite{Mallet2019Leveraging,Lim2018Molecular}.

In the field of molecule generation, GANs usually appear associated with RL. To fine-tune the generation of long SMILES strings, Guimaraes \etal \cite{Guimaraes2017Objective} employed a Wasserstein GAN \cite{Arjovsky2017Wasserstein} with a stochastic policy that increased the diversity, optimized the properties and maintained the drug-likeness of the generated samples. Sanchez-Langelin \etal \cite{Sanchez-Lengeling2017Optimizing} and Putin \etal \cite{Putin2018Reinforced} further improved upon this work to bias the distribution of generated molecules towards a goal. In addition, Mendez-Lucio \etal \cite{Mendez-Lucio2018De} used a GAN to generate molecules conditioned on gene expression signatures, which is particularly useful to create active compounds towards a certain target. Similarly to what is done with molecules, Killoran \etal \cite{Killoran2017Generating} employed a GAN to create realistic samples of DNA sequences from a small subset of configurations. The model was also tuned to design DNA chains adapted to protein binding and look for motifs representing functional roles. Adversarial training was also employed in the discovery of drugs for using molecular fingerprints as opposed to a reversible representation \cite{Kadurin2017druGAN,Kadurin2017cornucopia,Blaschke2018Application} and SMILES \cite{Polykovskiy2018Entangled}. However, avoiding the unstable training and mode collapse while generating molecules is still a hindrance for the usage of GANs in chemical design.

Although SMILES have proved to be a reliable representation for molecule generation, their sequential nature imposes some constraints to the architectures being learned. Forcing an RNN to implicitly learn their linguistic rules poses additional difficulties to the model under training. Additionally, decoding a sequence of generated characters into a valid molecule is especially difficult. In \cite{Gomez-Bombarelli2018Automatic}, the rate of success when decoding molecules depended on the proximity of the latent point to the valid molecule, and could be as low as 4\% for random points on the latent space. Although RL is as an alternative to reward the generation of valid molecules \cite{Jaques2017Sequence,Guimaraes2017Objective,Sanchez-Lengeling2017Optimizing}, other architecture changes can also circumvent this difficulty. Techniques to generate valid sequences imported from NLP studies include: using revision to improve the outcome of sequences \cite{Mueller2017Sequence}; adding a validator to the decoder to generate more valid samples \cite{Janet2018Accelerating}; introducing a grammar within the VAE to teach the model the fundamentals of SMILES strings \cite{Kusner2017Grammar}; using compiler theory to constrain the decoder to produce syntactically and semantically correct data \cite{Dai2018Syntax}; and using machine translation methods to convert between representations of sequences and/or grammar \cite{Winter2019Learning}.

Validity of generated sequences, however, is not the only thing that makes working with SMILES difficult. The sequential representation cannot represent similarity between molecules within edit distances \cite{Jin2018Junction} and a single molecule may have several different SMILES strings \cite{Bjerrum2017SMILES,Alperstein2019All}. The trade-off between processing this representation with text-based algorithms and discarding its chemical intuition calls for other approaches in the study and design of molecules.

\subsection{Molecular graphs}

An intuitive way of representing molecules is by means of its Lewis structure, computationally translated as a molecular graph. Given a graph $\mathcal{G} = \left(\mathcal{V}, \mathcal{E}\right)$, the atoms are represented as nodes $v_i \in \mathcal{V}$ and chemical bonds as edges $(v_i, v_j) \in \mathcal{E}$. Then, nodes and edges are decorated with labels indicating the atom type, bond type and so on. Many times, hydrogen atoms are treated implicitly for simplicity, since their presence can be inferred from traditional chemistry rules.

One of the first usages of graphs with DL for property prediction treated molecules as undirected cyclic graphs further processed using RNNs \cite{Lusci2013Deep}. Using graph convolutional networks \cite{Bruna2013Spectral}, Duvenaud \etal \cite{Duvenaud2015Convolutional} demonstrated the usage of machine-learned fingerprints to achieve better prediction of properties on neural networks. This approach started with a molecular graph and led to fixed-size fingerprints after several graph convolutions and a graph pooling layers. Kearnes \etal \cite{Kearnes2016Molecular} and Coley \etal \cite{Coley2017Convolutional} also evaluated the flexibility and promise of learned fingerprints from graph structures, especially because models could learn how to associate its chemical structure to their properties. Later, Gilmer \etal \cite{Gilmer2017Neural} unified graph convolutions as message-passing neural networks for quantum chemistry predictions, achieving DFT accuracy within their predictions of quantum properties, interpreting molecular 3D geometries as graphs with distance-labelled edges. Many more studies have explored the representative power of graphs within prediction tasks \cite{Hop2018Geometric,Yang2019Are}. These frameworks paved the way for using graph-based representations of molecules, especially because of their proximity with chemistry and geometrical interpretation.

\begin{figure}[tb]
\centering
\includegraphics[width=\linewidth]{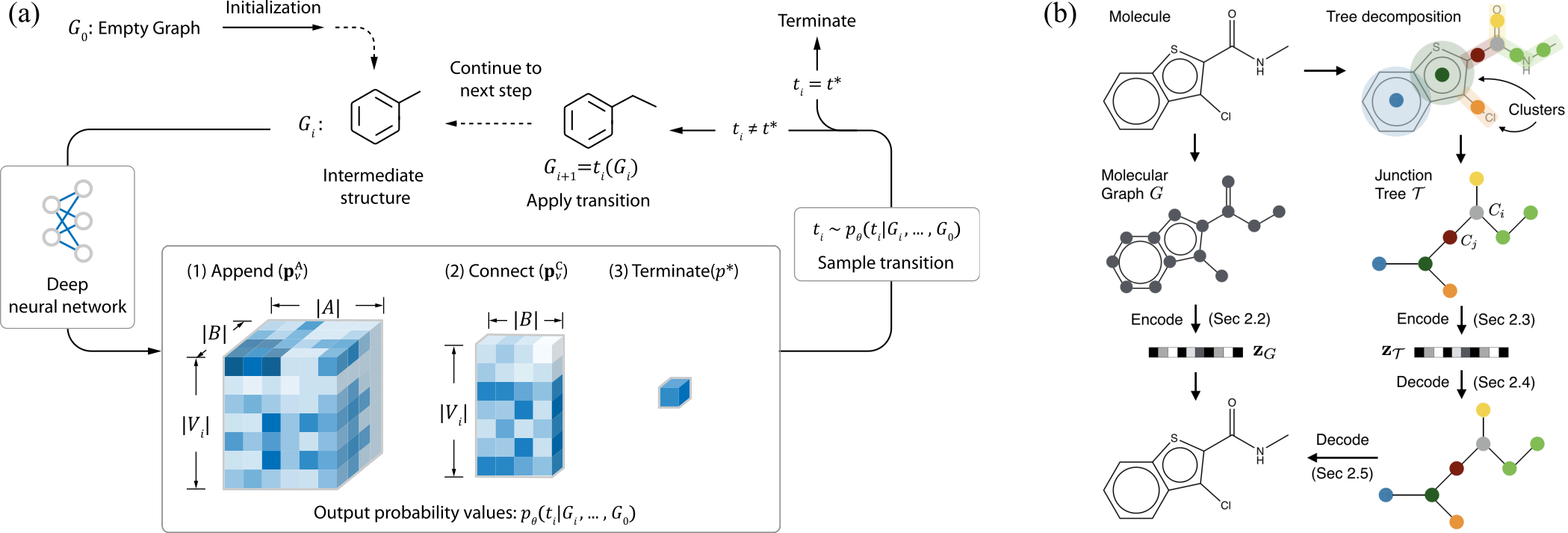}
\caption{Generative models for molecules using graphs. (a) Decision process for sequential generation of molecules from \cite{Li2018Multi}. (b) Junction Tree VAE for molecular graphs \cite{Jin2018Junction}. Figures reproduced from \cite{Li2018Multi,Jin2018Junction}.
\label{fig:graphs}}
\end{figure}

The generation of graphs is, however, non-trivial, especially because of the challenges imposed by graph isomorphism. As in SMILES strings, one way to generate molecular graphs is by sequentially adding nodes and edges to the graph. The sequential nature of decisions over graphs have already been implemented using an RNN \cite{You2018GraphRNN} for arbitrary graphs. Specifically for a small subset of graphs corresponding to valid molecules, Li \etal \cite{Li2018Multi} used a decoder policy to improve the outcomes of the model. The conditional generation of graphs allowed for molecules to be created with improved drug-likeness, synthetic accessibility, as well as allowed scaffold-based generations from a template (Fig. \ref{fig:graphs}a). Similar procedure was adopted by Li \etal \cite{Li2018Multi}, in which a graph-generating decision process using RNNs was proposed for molecules. These node-by-node generation rely on the ordering of nodes in the molecular graph and thus suffer with random permutations of the nodes.

In the VAE world, several methods have been proposed to deal with the problem of directly generating graphs from a latent code \cite{Kipf2016Variational,Simonovsky2018GraphVAE,Grover2018Graphite,Samanta2018Designing,Liu2018Constrained}. However, when working with reconstructions, the problem of graph isomorphism cannot be addressed without expensive calculations \cite{Simonovsky2018GraphVAE}. Furthermore, graph reconstructions suffer from validity and accuracy \cite{Simonovsky2018GraphVAE}, except when these constraints are enforced in the graph generation process \cite{Samanta2018Designing,Liu2018Constrained,Ma2018Constrained}. Currently, one of the most successful approaches to translate molecular graphs into a meaningful latent code while avoiding node-by-node generation is the Junction Tree Variational Auto-Encoder (JT-VAE) \cite{Jin2018Junction}. In this framework, the molecular graph is first decomposed into a vocabulary of subpieces extracted from the training set, which include rings, functional groups and atoms (see Fig. \ref{fig:graphs}b). Then, the model is trained to encode the full graph and the tree structure resulting from the decomposition into two latent spaces. A two-part reconstruction process is necessary to recover the original molecule from the two vector representations. Remarkably, the JT-VAE achieves 100\% of validity when generating small molecules, as well as 100\% of novelty when sampling the latent code from a prior. Moreover, a meaningful latent space is also seen for this method, which is essential for optimization and the automatic design of molecules. The authors later improve over the JT-VAE with graph-to-graph translation and auto-regressive methods towards molecular optimization tasks \cite{Jin2019Learning, Jin2019Multi}.

Other auto-regressive approaches combining VAE and sequential graph generation have been proposed to generate and optimize molecules. Assouel \etal \cite{Assouel2018DEFactor} introduced a decoding strategy to output arbitrarily large molecules based on their graph representation. The model, named DEFactor, is end-to-end differentiable, dispenses retraining during the optimization procedure and achieved high reconstruction accuracy ($>80\%)$ even for molecules with about 25 heavy atoms. Despite the restrictions on node permutations, DEFactor allows the direct optimization of the graph conditioned to properties of interest. This and other similar models also allow the generation of molecules based on given scaffolds \cite{Lim2019Scaffold}.

Auto-regressive methods for molecules have also been reported with the use of RL. Zhou \etal \cite{Zhou2018Optimization} created a Markov decision process to produce molecules with targeted properties through multi-objective RL. Similarly to what is done with graphs, this strategy adds bonds and atoms sequentially. However, as the actions are restricted to chemically valid ones, the model scores 100\% of validity in the generated compounds. The optimization process forgoes pre-training and allows flexibility in the choice of the importances for each objectives. As a follow-up to this work, the same group reports the usage of this generation scheme as a decoder in a RL-enhanced VAE for molecules \cite{Kearnes2019Decoding}.

In line with the usage of sequences of actions to create graphs, several groups have been working on different ways to represent and generate graphs through sequences. One approach is to split a graph in permutation-invariant N-gram path sets \cite{Liu2018N}, in analogy with NLP with atoms as words and molecules as sentences. This representation performs competitively with message-passing neural networks in classification and regression tasks. The combination of strings and graph methods is also seen in the work of Krenn \etal \cite{Krenn2019SELFIES}, which developed a sequence representation for general-purpose graphs. Their scheme shows high robustness against mutations in sequences and outperforms other representations (including SMILES strings) in terms of diversity, validity, and reconstruction accuracy when employed in sequence-based VAEs.

The adversarial generation of graphs is still very incipient, and few models of GANs with graphs have been demonstrated \cite{Guo2018Deep,Bojchevski2018NetGAN,Xiong2019DynGraphGAN}. De Cao and Kipf \cite{DeCao2018MolGAN} demonstrated MolGAN, a GAN trained with RL for generating molecular graphs, but their system is too prone to mode collapse. The output structure can be made discrete by differentiable processes such as Gumbel-softmax \cite{Jang2016Categorical,Kusner2016GANS}, but balancing the adversarial training with molecular constraints requires more study. P\"olsterl and Wachinger \cite{Poelsterl2019Likelihood} builds on MolGAN by adding an adversarial training to avoid calculating the reconstruction loss and extending the graph isomorphism network \cite{Xu2018How} to multigraphs. Further improvements include the approach from Maziarka \etal \cite{Maziarka2019Mol}, which relies on the latent space of a pretrained JT-VAE to produce and optimize molecules, and the work of Fan and Huang \cite{Fan2019Labeled}, which aims to generate labeled graphs.

While the combination of DL with graph theory and molecular design seems promising, large room for improvement is available in the field of graph generation. Outputting an arbitrary graph is still an open problem and scalability to larger graphs is still an issue for graphs \cite{Gilmer2017Neural}. Comparing graph isomorphism is a class-NP problem, and the measure of similarity between two graphs usually resort to expensive kernels or edit distances \cite{Neuhaus2007Bridging}, as are other problems with reconstruction, ordering and so on \cite{Li2018Learning}. In some cases, a distance metric can be defined for such data structures \cite{Schieber2017Quantification,Choi2018Comparing} or a set of networks can be trained to recognize similarity patterns within graphs \cite{Ktena2017Distance}. Furthermore, adding attention to graphs could also help in classification tasks \cite{Do2018Attentional} or in the extraction of structure-property relationships \cite{Ryu2018Deeply}, and specifying grammar rules for graph reconstruction may lead to improved results in molecular validity and stereochemistry \cite{Kajino2018Molecular}.

\section{Challenges and outlook for generative models}

The use of deep generative models is a powerful approach for teaching computers to observe and understand the real world. Far from being just a big-data crunching tool, DL algorithms can provide insights that augment human creativity \cite{Kalchbrenner2016Neural}. Completely evaluating a generative model is difficult \cite{Theis2015note}, since we lack an expression for the statistical distribution being learned. Nevertheless, by approximating real-life data with an appropriate representation, we are embedding intuition in the machine's understanding. In a sense, this is what we do, as human beings, when formulating theoretical concepts on chemistry, physics and many other fields of study. Furthering our limited ability to probe the inner workings of deep neural networks will allow us to transform learned embeddings into logical rules.

\begin{figure}[tb]
\centering
\includegraphics[width=\linewidth]{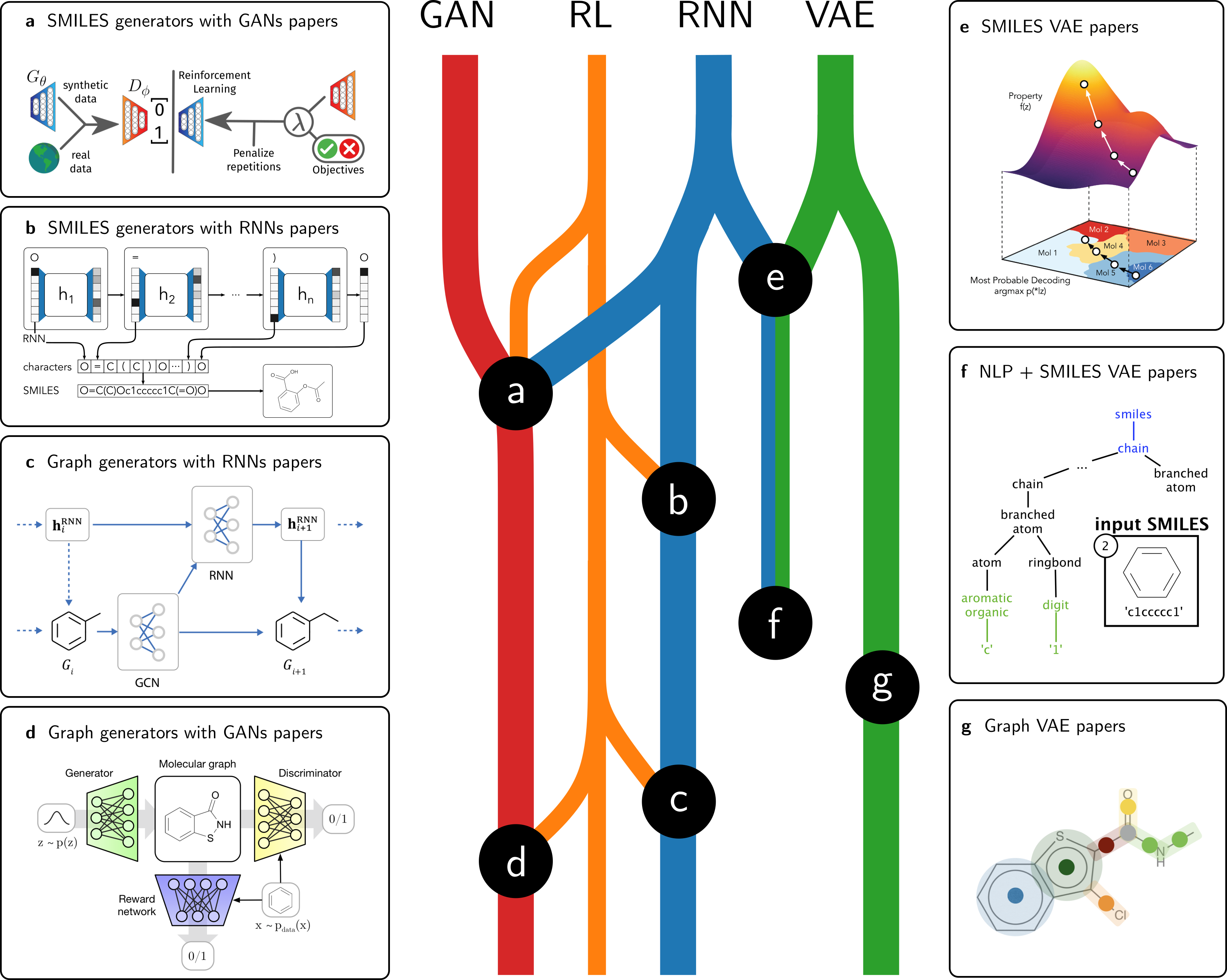}
\caption{Summary and timeline of current generative models for molecules. Newer models are located in the bottom of the diagram. Figures reproduced from \cite{Guimaraes2017Objective,Li2018Learning,DeCao2018MolGAN,Gomez-Bombarelli2018Automatic,Kusner2017Grammar,Jin2018Junction}.}
\label{fig:summary}
\end{figure}

In the field of chemical design, generative models are still in their infancy (see timeline summary in Fig. \ref{fig:summary}). While many achievements have been reported for such models, all of them share many challenges before a ``closed loop'' approach can be effectively implemented. Some of the trials are still inherent to all generative models: the generalization capability of a model, its power to make inferences on the real world, and capacity to bring novelty to it. In the chemical space, originality can be translated as the breadth and quality of possible molecules that the model can generate. To push forward the development of new technologies, we want our generative models to explore further regions of the chemical space in search of new solutions to current problems and extrapolate the training set, avoiding mode collapses or na\"ive interpolations. At the same time, we want it to capture rules inherent to the synthetically accessible space. Finally, we want to critically evaluate the performance of such models. Several benchmarks are being developed to assess the evolution of chemical generative models, providing quantitative comparisons beyond the mere prediction of solubility or drug-likeness \cite{Preuer2018Frechet,Polykovskiy2018Molecular,Wu2018MoleculeNet,Brown2019GuacaMol}.

The ease of navigation throughout the chemical space alone is not enough to determine a good model, however. Tailoring the generation of valid molecules for certain applications such as drug design \cite{Segler2018Generating} is also an important task. It reflects how well a generative model focus on the structure-property relationships for certain applications. This interpretation leads to even more powerful understandings of chemistry, and is closely tied to Gaussian processes \cite{Gomez-Bombarelli2018Automatic}, Bayesian optimization \cite{Haese2018PHOENICS}, and virtual screening.

In the generation process, outputting an arbitrary molecule is still an open problem and is closely conditioned to the representation. While SMILES have been demonstrated useful to represent molecules, graphs are able to convey real chemical features in it, which is useful for learning properties from structures. However, three-dimensional atomic coordinates should be considered for decoding as well. Recent works are going well beyond the connectivity of a molecule to provide equilibrium geometries of molecules using generative models \cite{Gebauer2018Generating,Noe2018Boltzmann,Gebauer2019Symmetry,Joergensen2019Atomistic,Mansimov2019Molecular}. This is crucial to bypass expensive sampling of low-energy configurations from the potential energy surface of molecules. We should expect advances not only on decoding and generating graphs from latent codes, but also in invertible molecular representations in terms of sequences, connectivity and spatial arrangement.

Finally, as the field of generative models advances, we should expect even more exciting models to design molecules. The normalizing-flow based Boltzmann Generator \cite{Noe2018Boltzmann} and GraphNVP \cite{Madhawa2019GraphNVP} are examples of models based on more recent strategies. Furthermore, the use of generative models to understand molecules in an unsupervised way advances along with the inverse design, from coarse-graining \cite{Wang2018Machine,Wang2018Coarse} and synthesizability of small molecules \cite{Bradshaw2019Generative,Bradshaw2019Model} to genetic variation in complex biomolecules \cite{Riesselman2018Deep}.

In summary, generative models hold promise to revolutionize the chemical design. Not only they allow optimizations or learn directly from data, but also bypass the necessity of a human supervising the generation of materials. Facing the challenges among these models is essential for accelerating the discovery cycle of new materials and, perhaps, improvement of the human understanding of the nature.

\begin{acknowledgement}
D.S.-K. acknowledges the MIT Nicole and Ingo Wender Fellowship and the MIT Robert Rose Presidential Fellowship for financial support. R.G.-B. thanks MIT DMSE and Toyota Faculty Chair for support.
\end{acknowledgement}

\bibliographystyle{spphys}
\bibliography{references}

\end{document}